\documentclass[a4paper,twoside]{article}

\usepackage{epsfig}
\usepackage{subcaption}
\usepackage{calc}
\usepackage{amssymb}
\usepackage{amstext}
\usepackage{amsmath}
\usepackage{amsthm}
\usepackage{multicol}
\usepackage{pslatex}
\usepackage[round]{natbib} 
\usepackage{algorithm2e}
\usepackage[bottom]{footmisc}
\usepackage{array} 
\usepackage{booktabs}
\newcolumntype{P}[1]{>{\centering\arraybackslash}p{#1}}

\usepackage{SCITEPRESS}     

\DeclareMathOperator*{\argmax}{arg\,max}

\begin{document}

\title{Proposing Hierarchical Goal-Conditioned Policy Planning\\ in Multi-Goal Reinforcement Learning
\thanks{This is the preprint version of the paper accepted at ICAART 2025.}
}

\author{Gavin B.\ Rens\\
Computer Science Division, Stellenbosch University, Stellenbosch, South Africa\\
gavinrens@sun.ac.za\\[5mm]
December 2024}


\abstract{Humanoid robots must master numerous tasks with sparse rewards, posing a challenge for reinforcement learning (RL). We propose a method combining RL and automated planning to address this. Our approach uses short goal-conditioned policies (GCPs) organized hierarchically, with Monte Carlo Tree Search (MCTS) planning using high-level actions (HLAs). Instead of primitive actions, the planning process generates HLAs. A single plan-tree, maintained during the agent's lifetime, holds knowledge about goal achievement. This hierarchy enhances sample efficiency and speeds up reasoning by reusing HLAs and anticipating future actions. Our Hierarchical Goal-Conditioned Policy Planning (HGCPP) framework uniquely integrates GCPs, MCTS, and hierarchical RL, potentially improving exploration and planning in complex tasks.}

\onecolumn \maketitle \normalsize \setcounter{footnote}{0} \vfill

\section{\uppercase{Introduction}}
\label{sec:introduction}
Humanoid robots have to learn to perform several, if not hundreds of tasks. For instance, a single robot working in a house will be expected to pack and unpack the dishwasher, pack and unpack the washing machine, make tea and coffee, fetch items on demand, tidy up a room, etc. For a reinforcement learning (RL) agent to discover good policies or action sequences when the tasks produce relatively sparse rewards is challenging \citep{prefjl20,ehlsc21,sk23,hwhw23,lwjz23}.
Except for \citep{ehlsc21}, the other four references use hierarchical approaches.
This paper proposes an approach for agents to learn multiple tasks (goals) drawing from techniques in hierarchical RL and automated planning.

A high-level description of our approach follows.
The agent learns short goal-conditioned policies which are organized into a hierarchical structure. Monte Carlo Tree Search (MCTS) is then used to plan to complete all the tasks. Typical actions in MCTS are replaced by high-level actions (HLAs) from the hierarchical structure. The lowest-level kind of HLA is a goal-conditioned policy (GCP). Higher-level HLAs are composed of lower-level HLAs. Actions in \textit{our} version of MCTS can be any HLAs at any level. We assume that the primitive actions making up GCPs are given/known. But the planning process does not operate directly on primitive actions; it involves generating HLAs \textit{during} planning.

\begin{figure}
    \centering
    \includegraphics[width=1.0\linewidth]{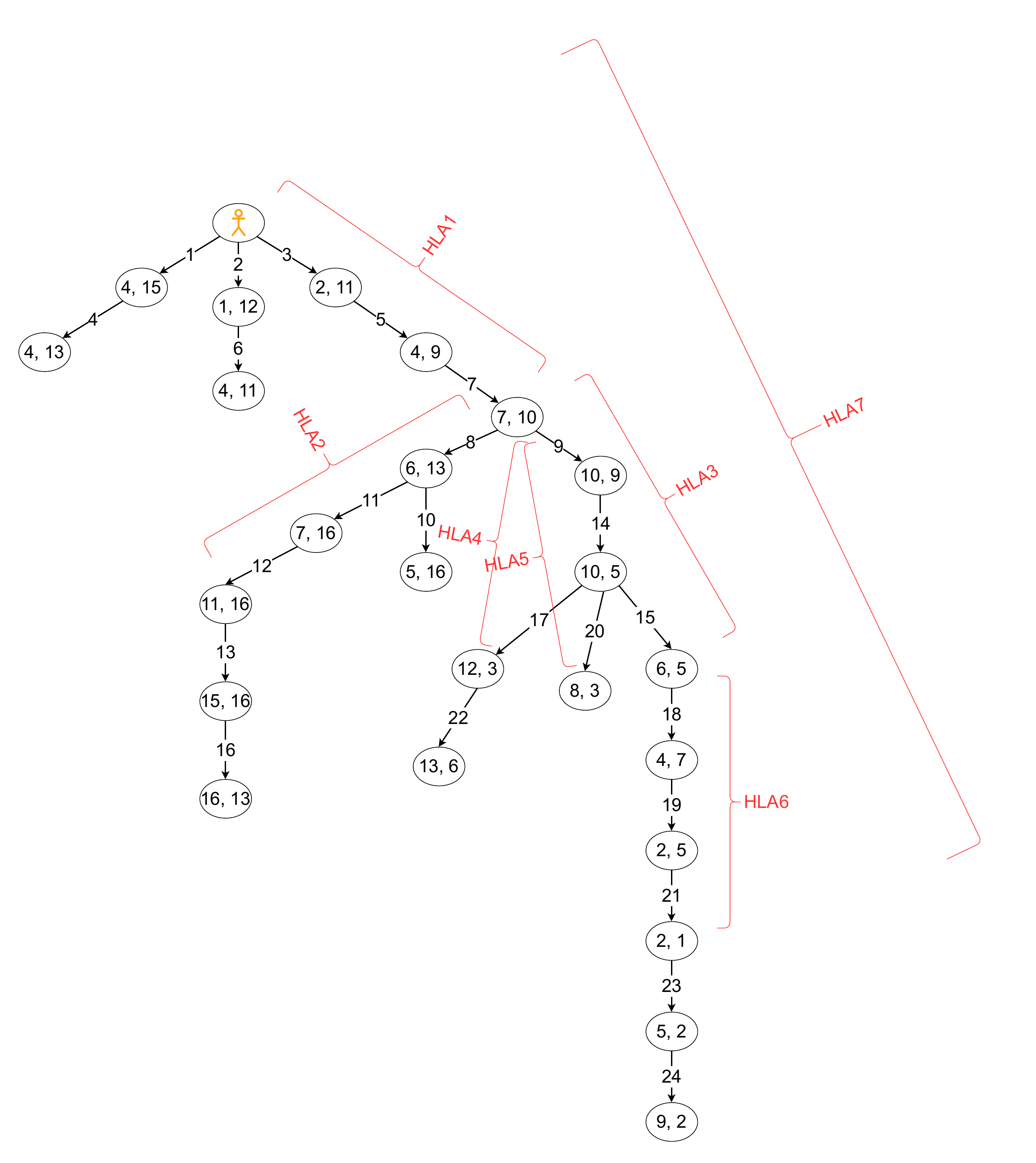}
    \caption{Complete plan-
tree corresponding to the maze grid-world. Note that HLA7
is composed of HLA1, HLA3 and HLA6, in that order.}
    \label{fig:final-plan-tree}
\end{figure}

A single plan-tree is grown and maintained during the lifetime of the agent. The tree constitutes the agent's knowledge about how to achieve its goals (how to complete its tasks). Figure~\ref{fig:final-plan-tree} is a complete plan-tree for the environment depicted in Figure~\ref{fig:maze-grid-world}. The idea is to associate a value for each goal with every HLA (at every level) discovered by the agent so far. An agent becomes more sample efficient by reusing some of the same HLAs for reaching different goals. Moreover, the agent can reason (search) faster by looking farther into the future to find more valuable sequences of actions than if it considered only primitive actions. This is the conventional reason for employing hierarchical planning; see the papers referenced in Section \ref{sec:HRL} and Section \ref{sec:Related-Work}

For ease of reference, we call our new approach HGCPP (Hierarchical Goal-Conditioned Policy Planning).
To the best of our knowledge, at the time of writing, no-one has explicitly combined goal-conditioned policies, MCTS, and hierarchical RL in a single framework.
This combination could potentially lead to more efficient exploration and planning in complex domains with multiple long-horizon goals. 

The rest of the paper is organized as follows.
Section \ref{sec:Background} provides the necessary background theory.
Section \ref{sec:Related-Work} reviews the related work.
Section \ref{sec:The-Algo} presents our framework (or family of algorithms). We also propose some novel and existing techniques that could be used for implementing an instance of HGCPP.
In Section \ref{sec:Discussion} we further analyze our proposed approach and make further suggestions to improve it.
As this is early-stage research, there is no evaluation yet.

\section{\uppercase{Background}}\label{sec:Background}

\subsection{Goal-Conditioned Reinforcement Learning}

Reinforcement learning (RL) is based on the Markov decision process (MDP). An MDP is described as a tuple $\langle S,A,T,R,\gamma \rangle$, where $S$ is a set of states, $A$ is a set of (primitive) actions, $T$ is a transition function (the probability of reaching a state from a state via an action), $R:S\times A\times S\mapsto \mathbb{R}$ and $\gamma$ is the discount factor.
The value of a state $s$ is the expected total discounted reward an agent will get from $s$ onward, that is,
\[
V(s)\doteq \mathbb{E}_{s_{t+1}\sim T(s_t,a_t,\cdot)}[\sum_{t=0}^\infty \gamma^t R(s_t,a_t,s_{t+1})\mid s_0=s],
\]
where $s_{t+1}$ is the state reached by executing action $a$ in state $s_t$.
Similarly, the value of executing $a$ in $s$ is defined by 
\[
Q(a,s)\doteq R(s,a,s') + \gamma\sum_{s'\in S}T(s,a,s') V(s').
\]
We call it the Q function and its value is a q-value.
The aim in MDPs and RL is to maximize $V(s)$ for all reachable states $s$.
A policy $\pi:S\mapsto A$ tells the agent what to do: execute $a=\pi(s)$ when in $s$.
A policy can be defined in terms of a Q function:
\[
\forall s \in S, \pi(s)\doteq \argmax_{a\in A}Q(a,s).
\]
It is thus desirable to find `good' q-values (see later).

Goal-conditioned reinforcement learning (GCRL) \citep{shgs15,lzz22} is based on the GCMDP, defined as the tuple $\langle S,A,T,R,G,\gamma,\rangle$, where $S$, $A$, $T$ and $\gamma$ are as before, $G$ is a set of (desired) goals and $R:S\times A\times S\times G\mapsto \mathbb{R}$ is goal-conditioned reward function.
In GCRL, the value of a state and the Q function are conditioned on a goal $g\in G$: $V(s,g)$, respectively, $Q(a,s,g)$. 
A goal-conditioned policy is
\[
\pi:S\times G\mapsto A
\]
such that $\pi(s,g)$ is the action to execute in state $s$ towards achieving $g$.

\subsection{Hierarchical Reinforcement Learning}\label{sec:HRL}

Traditionally, hierarchical RL (HRL) has been a divide-and-conquer approach, that is, determine the end-goal, divide it into a set of subgoals, then select or learn the best way to reach the subgoals, and finally, achieve each subgoal in an order appropriate to reach the end-goal.

``Hierarchical Reinforcement Learning (HRL) rests on finding good re-usable temporally extended actions that may also provide opportunities for state abstraction. Methods for reinforcement learning can be extended to work with abstract states and actions over a hierarchy of subtasks that decompose the original problem, potentially reducing its computational complexity.'' \citep{h12}

``The hierarchical approach has three challenges [619, 435]: find subgoals, find a meta-policy over these subgoals, and find subpolicies for these subgoals.'' \citep{p23}

The divide-and-conquer approach can be thought of as a top-down approach. The approach that our framework takes is bottom-up: the agent learns `skills' that could be employed for achieving different end-goals, then memorizes sets of connected skills as more complex skills. Even more complex skills may be memorized based on less complex skills, and so on. Higher-level skills are always based on already-learned skills. In this work, we call a skill (of any complexity) a \textit{high-level action}.

\subsection{Monte Carlo Tree Search}

The version Monte Carlo tree search (MCTS) we are interested in is applicable to single agents based on MDPs \citep{ks06}.

An MCTS-based agent in state $s$ that wants to select its next action loops thru four phases to generate a search tree rooted at a node representing $s$. Once the agent's planning budget is depleted, it selects the action extending from the root that was most visited (see below) or has the highest q-value. While there still is planning budget, the algorithm loops thru the following phases \citep{bpwlctpsc12}.

\textbf{Selection}
A route from the root until a leaf node is chosen.\footnote{For discrete problems with not `too many' actions, a node is a leaf if not all actions have been tried at least once in that node.}
Let $UCBX:Nodes\times A\mapsto\mathbb{R}$ denote any variant of the Upper Confidence Bound, like UCB1 \citep{ks06}.
For every non-leaf node $n$ in the tree, follow the path via action $a^*=\argmax_{a\in A}UCBX(n,a)$.

\textbf{Expansion}
When a leaf node $n$ is encountered, select an action $a$ not yet tried in $n$ and generate a new node representing $s'$ as child of $n$.

\textbf{Rollout}
To estimate a value for $s'$ (or $n'$ representing it), simulate one or more Monte Carlo trajectories (rollouts) and use them to calculate a reward-to-go from $n'$. If the rollout value for $s'$ is $V(s')$, the $Q(a,s)$ can be calculated.

\textbf{Backpropagation}
If $n'$ has just been created and the q-value associated with the action leading to it has been determined, then all action branches on the path from the root till $n$ must be updated. That is, the change in value at the end of a path must be propagated bach up the tree.

In this paper, when we write $f(n)$ (where $n$ might have indices and $f$ is some function), we generally mean $f(n.s)$ and $f(s')$, where $n.s=s'$ is the state represented by node $n$.

\subsection{Multi-Objective/Goal Reinforcement Learning}

We make a distinction between multi-objective RL (MORL) and multi-goal RL (MGRL).
MORL \citep{lxh15} attempts to achieve all goals simultaneously. There is often a weight or priority assigned to goals.
MGRL \citep{k93,smddpwp11} does not weight any goal as more or less important and each goal is assumed to eventually be pursued individually. There is no question about which goal is more important; the agent simply pursues the goal it is commanded to.
In both cases, the agent can learn all goals simultaneously (broadly speaking).
When MORL goals are prioritized and MGRL has a curriculum, the two frameworks become very similar.


Our framework follows the MGRL approach, not the MORL approach.

\section{\uppercase{Related Work}}\label{sec:Related-Work}

The areas of hierarchical reinforcement learning (HRL) and goal-conditioned reinforcement learning (GCRL) are very large. The overlapping area of goal-conditioned hierarchical RL is also quite large. Table~\ref{tab:rw} shows where some of the related work falls with respect to the dimensions of 1) goal-conditioned (GC), 2) hierarchical (H), 3) planning (P) and 4) RL.

\begin{table}[h!]
\centering
\begin{small}
\begin{tabular}{|l|P{15mm}P{15mm}P{15mm}|}
\hline
 & only RL & RL+P & only P \\
\hline
only GC & 12, 3, 25, 22, 2, 6 & 4, 15 & \\
GC+H & 9, 17, 10, 18, 19, 20 & 1, 5, 11, 13, 14, 16, 23, 24, 26, HGCPP & 7, 21\\
only H &  &  & 8\\
\hline
\end{tabular}
\end{small}
\caption{Related work categorized over four dimensions. The numbers map to the literature in the list below.}
\label{tab:rw}
\end{table}

\begin{tabular}{p{35mm}p{35mm}}
\begin{scriptsize}
\begin{itemize}
    \itemsep=0pt
    \item[1] \citep{knst16} 
    \item[2] \citep{awrsfwmtaz17} 
    \item[3] \citep{fhga18}
    \item[4] \citep{npll19}
    \item[5] \citep{pc19}
    \item[6] \citep{fzdhz19}
    \item[7] \citep{prefjl20}
    \item[8] \citep{lzgjc20}
    \item[9] \citep{csl23}
    \item[10] \citep{kss21}
    \item[11] \citep{wsv21}
    \item[12] \citep{ehlsc21}
    \item[13] \citep{fynl22}
\end{itemize}
\end{scriptsize}
    &
\begin{scriptsize}
\begin{itemize}
    \itemsep=0pt
    \item[14] \citep{lttz22}
    \item[15] \citep{msbla22}
    \item[16] \citep{yjwlwls22}
    \item[17] \citep{pgel23}
    \item[18] \citep{sk23}
    \item[19] \citep{zmn23}
    \item[20] \citep{hwhw23}
    \item[21] \citep{lwjz23}
    \item[22] \citep{csl23}
    \item[23] \citep{wwykp24}
    \item[24] \citep{ljslzjh24}
    \item[25] \citep{wjm24}
    \item[26] \citep{blwogst24}
\end{itemize}
\end{scriptsize}
\end{tabular}

We do not have the space for a full literature survey of work involving all combinations of two or more of these four dimensions. Nonetheless, we believe that the work mentioned in this section is fairly representative of the state of the art relating to HGCPP.


We found only two papers involving hierarchical MCTS: \citep{lzgjc20} does not involve RL nor goal-conditioning, nor multiple-goals. It is placed in the bottom right-hand corner of Table~\ref{tab:rw}.

The work of \cite{pc19} is the closest to our framework.
They proposes a method for HRL in fighting games using \textit{options} with MCTS. Instead of using all possible game actions, they create subsets of actions (options) for specific gameplay behaviors. Each option limits MCTS to search only relevant actions, making the search more efficient and precise. Options can be activated from any state and terminate after one step, allowing the agent to learn when to use and switch between different options. The approach uses Q-learning with linear regression to approximate option values, and an $\epsilon$-greedy policy for option selection.
They call their approach \textit{hierarchical} simply because it uses options. HGCPP has hierarchies of arbitrary depth, whereas theirs is flat. They do not generate subgoals, while HGCPP generates reachable and novel behavioral goals to promote exploration. One could view their approach as being implicitly multi-goal.
\cite{pc19} is in the group together with HGCPP in Table~\ref{tab:rw}.

\section{\uppercase{The Proposed Algorithm}}\label{sec:The-Algo}

We first give an overview, then give the pseudo-code. Then we discuss the novel and unconventional concepts in more detail.

We make two assumptions--that
\begin{itemize}
    \item the agent will always start a task from a specific, pre-defined state (location, battery-level, etc.) and
    \item the agent may be given some subgoals to assist it in learning how to complete a task.
\end{itemize}
Future work could investigate methods to deal with task environments where these assumptions are relaxed.

We define a \textit{contextual goal-conditioned policy} (CGCP) as a policy parameterized by a context $C\subset S$ and a (behavioral) goal $g\subset S$, where $S$ is the state-space. The context gives the region from which the goal will be pursued. In this paper, we simplify the discussion by equating $C$ with one state $s_s$ and by equating $g$ with one state $s_e$. A CGCP is something between an \textit{option} \citep{sps99} and a GCP.
Formally,
\[\pi: C\times S \times S\mapsto A \hspace{4mm} \mbox{ or} \hspace{5mm} \pi[s_s,s_e] : S \mapsto A\]
is a CGCP. The idea is that $\pi[s_s,s_e]$ can be used to guide an agent from $s_s$ to $s_e$.
Whereas GCPs in traditional GCRL do not specify a context, we include it so that one can reason about all policies relevant to a given context. In the rest of this paper we refer to CGCPs simply as GCPs.

Two GCPs, $\pi[C,g]$ as predecessor and $\pi'[C',g']$ as successor, are \textit{linked} if and only if $g\cap C'\neq \emptyset$. When $g=\{s_e\}$ and $C'=\{s'_s\}$, then we require that $s_e=s'_s$ for $\pi$ and $\pi'$ to be linked.
We represent that $\pi$ is linked to $\pi'$ as $\pi\to\pi'$.

Assume that the agent has already generated some/many HLAs during MCTS.
Let $H$ denote the set of all HLAs generated so far. 
They are organized in a tree structure (which we shall call a \textit{plan-tree}). Each HLA can be decomposed into several lower-level HLAs, unless the HLA is a GCP. To indicate the start and end states of an HLA $h$, we write $h[s_s,s_e]$. ``Assume $h[s_s,s_e]$.'' is redd `Assume $h$ is an HLA starting in state $s_s$ and ending in state $s_e$'.
Notation $HLAs(s)$ refers to all HLAs starting in state $s$, or $HLAs(n)$ refers to all HLAs starting in node $n$ (leading to children of $n$).

The exploration-exploitation strategy is determined by a function $expand$.
$R^\pi$ is the value of a (learned) GCP $\pi$.
Every time a new GCP is leanrt, all non-GCP HLAs on the path from the root node $n_{root}$ to the GCP must be updated. This happens after/due to the backpropagation of $R^\pi$ to all linked GCPs on the path from the root.

If GCP $\pi[s_s,s_e]$ is the parent of GCP $\pi'[s'_s,s'_e]$ in the plan-tree, then $\pi\to\pi'$. Suppose $\pi[s_s,s_e]$ is an HLA of node $n$ in the plan-tree (s.t.\ $n.s=s_s$), then there should exist a node $n'$ which is a child of $n$ such that $n'.s=s_e$. If $\pi\to\pi'$, then it means that $\pi'[s'_s,s'_e]$ is an HLA of node $n'$ such that $s_e=s'_s$.
We define linked GCPs of arbitrary length as 
\[\pi_0\to\pi_1\to \cdots \to\pi_m \mbox{ for } m>0.\]

\begin{figure}
    \centering
    \includegraphics[scale=0.24]{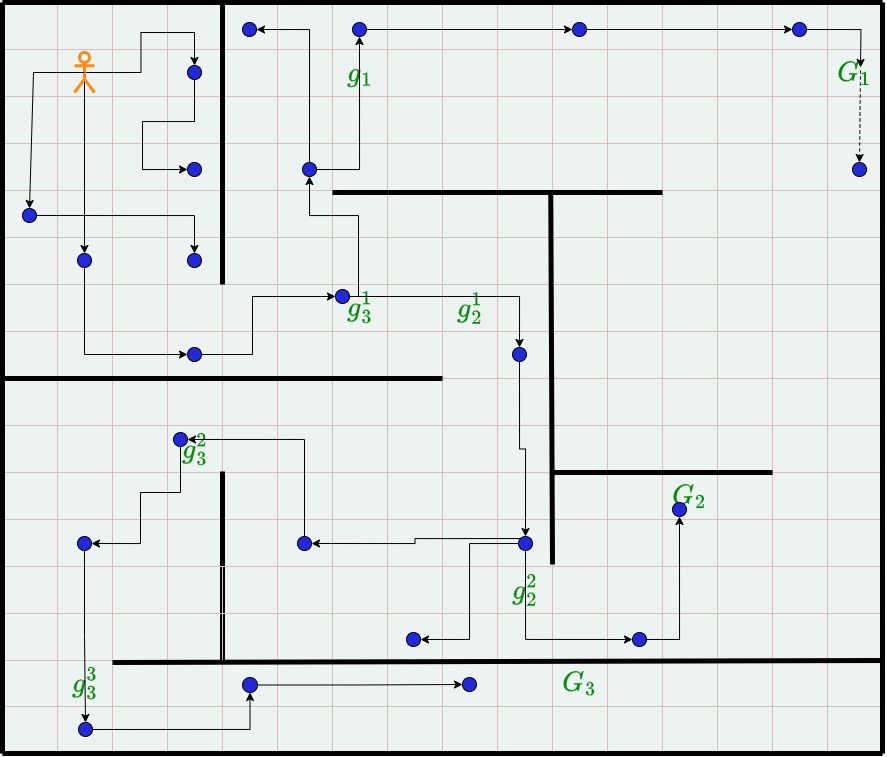}
    \caption{Maze grid-world with three main desired goals, $G_1$, $G_2$ and $G_3$, and their waypoints as desired sub-goals. Blue dots indicate endpoints of GCPs; blue dots are also behavioral goals. Arrows show typical trajectories of GCPs.}
    \label{fig:maze-grid-world}
\end{figure}

\begin{figure*}
    \centering
    \includegraphics[scale=0.2]{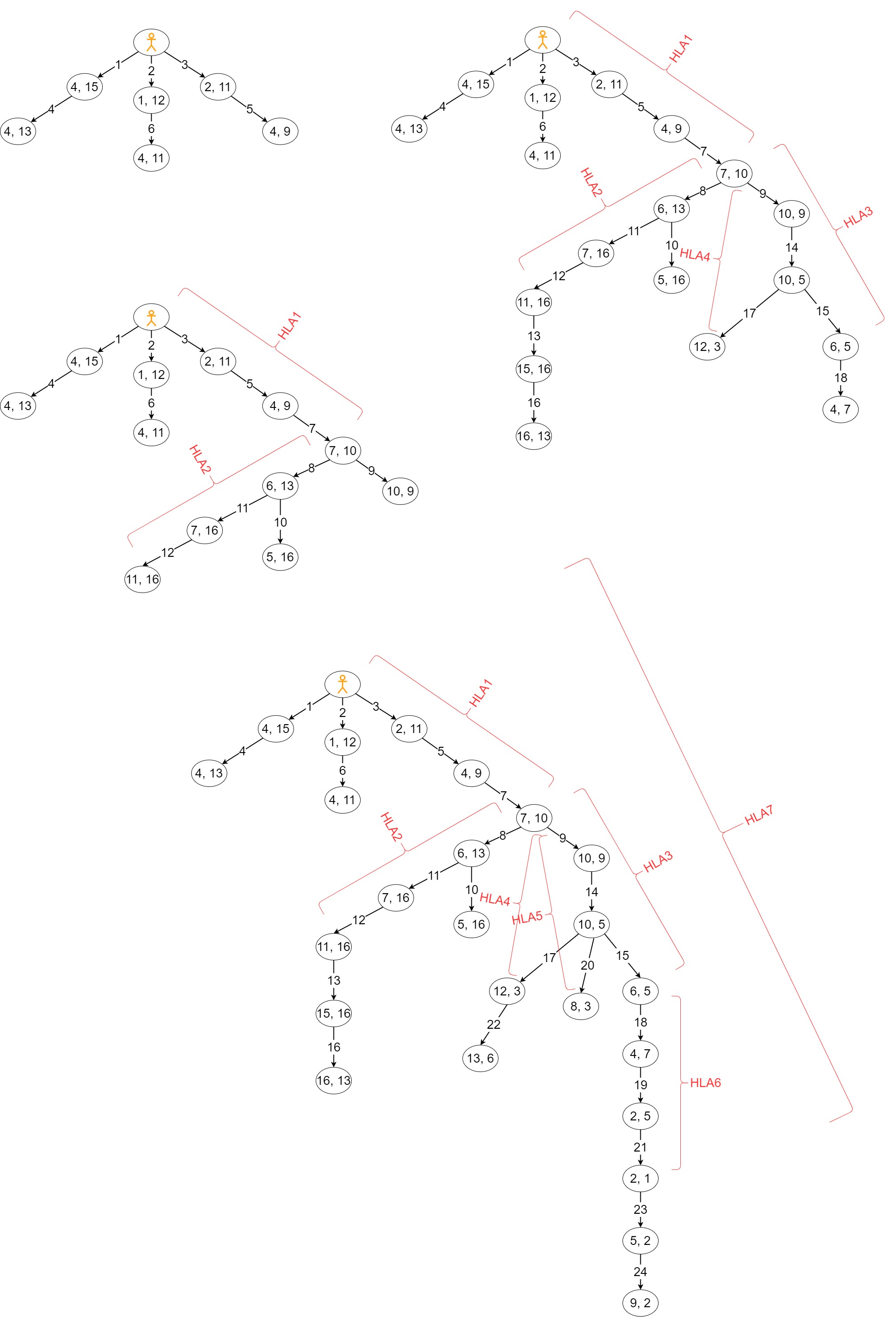}
    \caption{Four successive plan-trees. Top left: plan-tree after generating six GCPs. Middle left: plan-tree after generating twelve GCPs. Top right: plan-tree after generating eighteen GCPs. The two larger diagrams show which linked GCPs form higher-level HLAs. Bottom: Complete plan-tree corresponding to the maze grid-world. Note that HLA7 is composed of HLA1, HLA3 and HLA6, in that order.}
    \label{fig:plan-tree}
\end{figure*}

To illustrate some of the ideas mentioned above, look at Figures \ref{fig:maze-grid-world} and \ref{fig:plan-tree}.
Figures \ref{fig:maze-grid-world} shows a maze grid-world where an agent must learn sequences of HLAs to achieve three desired (main) goals. The other figure shows various stages of the plan-tree as it is being grown. The numbers labeling the arrows indicate the order in which the GCPs are generated.

In this work, we take the following approach regarding HLA generation. A newly generated HLA is always a GCP. After a number of linked GCPs (say three) have been generated/learned, they are composed into an HLA at one level higher. And in general, a sequence of $n$ HLAs at level $\ell$ are composed into an HLA at level $\ell + 1$. Of course, some HLAs are at the highest level and do not form part of a higher-level HLA.
This approach has the advantage that every behavioral goal at the end of an HLA is reachable if the constituent GCPs exist.
The HGCPP algorithm is as follows.

Initialize the root node of the MCTS tree: $n_{root}$ such that $n_{root}.s = s_{init}$.
Initialize $g^*\in G$ current desired goal to focus on.
\begin{itemize}
    \item[1.] $select\gets \mathit{false}$, \\$select \gets true \sim expand(x,n,\delta)$\hfill \# see \S~\ref{sec:expansion-strat}
    \item[2.] If $select$:\hfill \# exploit
    \begin{itemize}
        \item[a.] $h^* \gets \max_{h\in HLAs(n)}UCBX(n,h,g^*)$
        \item[b.] $n\gets n'$ s.t.\ $n'.s = s_e$ and $h^*[s_s,s_e]$
        \item[c.] Goto step 1
    \end{itemize}
    \item[3.] If not $select$: \hfill \# explore
    \begin{itemize}
        \item[a.] $s'\gets ChooseBevGoal(n)$ \hfill \# see \S~\ref{sec:Sampling-Behavioral-Goals}
        \item[b.] Generate $n'$ s.t.\ $n'.s = s'$ and add as child of $n$
        \item[c.] Attempt to learn $\pi[n,s']$
        \item[d.] Add $\pi[n,s'']$ to $HLAs(n)$ \hfill \# see \S~\ref{sec:Opportunism}
        \item[e.] For every $g\in G$, initialize $Q(\pi,n,g)\gets R^{\pi[n,s'']}_g + Rollout_g(s'',1)$\hfill \# see \S~\ref{sec:GCP-value} \& \ref{sec:Updating-of-HLAs}
        \item[f.] For every $g\in G$, backpropagate $Q(\pi,n,g)$\\ \phantom{a} \hfill \# see \S~\ref{sec:backprop} 
        \item[g.] Create all new HLAs $h$ as applicable: $h=\pi_0\to\cdots\to\pi_m$, where $\pi_0=\pi[n^0,\cdot]$ and $\pi_m=\pi[n,s'']$
        \item[h.] Add all $h$ to $HLAs(n^0)$
        \item[i.] Update every affected HLA \hfill \# see \S~\ref{sec:Updating-of-HLAs}
    \end{itemize}
    \item[5.] $g^*\gets \mathit{focus}(g^*,G)$\hfill \# see \S~\ref{sec:focus}
    \item[6.] $n\gets n_{root}$
    \item[7.] Goto step 1
\end{itemize}

\subsection{Sampling behavioral goals}\label{sec:Sampling-Behavioral-Goals}

If the agent decides to explore and thus to generate a new GCP, what should the end-point of the new GCP be? That is, if the agent is at $s$, how does the agent select $s'$ to learn $\pi[s, s']$? 
The `quality' of a behavioral goal $s'$ to be achieved from current exploration point $s$ is likely to be based on curiosity \citep{s06,s91} and/or novelty-seeking \citep{ls11,cmslsc17} and/or coverage \citep{va06,hls19} and/or reachability \citep{csl23}.

We require a function that takes as argument the exploration point $s$ and maps to behavioral goal $s'$ that is similar enough to $s$ that it has good reachability (not too easy or too hard to achieve it, aka a GOID - goal of intermediate difficulty \citep{csl23}). Moreover, of all $s''$ with approximately equal similarity to $s$, $s'$ must be the $s''$ most dissimilar to all children of $s$ on average. The latter property promotes novelty and (local) coverage.

We denote the function that selects the `appropriate' behavioral goal when at exploration point $s$ as $ChooseBevGoal(s)$.
Some algorithms have been proposed in which variational autoencoders are trained to represent $ChooseBevGoal(s)$ with the desired properties \citep{knjl21,fynl22,lttz22}: An encoder represents states in a latent space, and a decoder generates a state with similarity proportional the (hyper)parameter.
They sample from the latent space and select the most applicable subgoals for the planning phase \citep{knjl21,fynl22} or perturb the subgoal to promote novelty \citep{lttz22}.
Another candidate for representing $ChooseBevGoal(s)$ is the method proposed by \citet{csl23}: use a set of particles updated via Stein Variational Gradient Descent ``to fit GOIDs, modeled as goals whose success is the most unpredictable.''
There are several other works that train a neural network that can then be used to sample a good goal from any exploration point \citep[e.g.]{sk23,wwykp24}.

In HGCPP, each time the node representing $s$ is expanded, a new behavioral goal $g_b$ is generated. Each new $g_b$ to be achieved from $s$ must be as novel as possible given the behavioral goal already associated with (i.e.\ connecting the children of) $s$. We thus propose the following. Sample a small batch $B$ of candidate behavioral goals using a pre-trained variational autoencoder and select $g_b\in B$ that is most different to all existing behavioral goals already associated with $s$. The choice of measure of difference/similarity between two goals is left up to the person implementing the algorithm.

\subsection{The expansion strategy}\label{sec:expansion-strat}

For every iteration thru the MCTS tree, for every internal node on the search path, a decision must be made whether to follow one of the generated HLAs or to explore and thus expand the tree with a new HLA. 

Let $\eta(s,\delta)$ be an estimate for the number candidate behavioral goals $g_b$ around $s$, with $\delta$ being a hyper-parameter proportional to the distance of $g_b$ from $s$.
We propose the following exploration strategy, based on the logistic function.
Expand node $n$ with a probability equal to 
\begin{equation}
    expand(x,n,\delta) = \frac{1}{1+e^{k(\eta(n,\delta)/2-x)}},
\end{equation}
where $0<k\leq 1$ and $x$ is the number of GCPs starting in $n$.
If we do not expand, then we select an HLA from $HLAs(n)$ that maximizes $UCBX$.

\subsection{The value of a GCP}\label{sec:GCP-value}

Suppose that we are learning policy $\pi[s_s,s_e]$.
Let $\sigma[s_s,s_e]$ be a sequence $s_1, a_1, s_2, a_2,\ldots, a_j, s_{j+1}$ of states and primitive actions such that $s_1=s_s$ and $s_{j+1}=s_e$.
Let $\mathit{traj}(s_s,s_e)$ be all such trajectories (sequences) $\sigma[s_s,s_e]$ experienced by the agent in its lifetime.
Then we define the value $R^{\pi[s_s,s_e]}_g$ of $\pi[s_s,s_e]$ with respect to goal $g$ as
\begin{equation*}
     \frac{1}{\vert \mathit{traj}(s_s,s_e)\vert}\sum_{\sigma[s_s,s_e]\in \mathit{traj}(s_s,s_e)}\sum_{s_i,a_i\in\sigma[s_s,s_e]}R_g(a_i,s_i).
\end{equation*}
In words, $R^{\pi[s_s,s_e]}_g$ is the average sum of rewards experienced of actions executes in trajectories from $s_s$ to $s_e$ for pursuing $g$.
At the moment, we do not specify whether $R_g(a_i,s_i)$ is given or learned.

\subsection{Backpropagation}\label{sec:backprop}

Only GCPs are directly involved in backpropagation. In other words, we backpropagate values from node to node up the plan-tree as usual in MCTS while treating the GCPs as primitive actions in regular MCTS. The way the hierarchy is constructed guarantees that there is a sequence of linked GCPs between any node reached and the root node.

Every time a GCP $\pi[n.s,n'.s]$ and corresponding node $n'$ are added to the plan-tree, the value of $\pi$ determined just after learning $\pi$ (i.e. $R^\pi_g$) is propagated back up the tree, for every desired goal.
In general, for every GCP $\pi[n.s,\cdot]$ (and every non-leaf node $n$) on the path from the root until $n'$, for every goal $g\in G$, we maintain a goal-conditioned value $Q(\pi,n,g)$ representing the estimated reward-to-go from $n.s$ onwards.

As a procedure, backpropagation is defined by Algorithm~\ref{alg:bp}.
Note that $q$ and $fv$ are arrays indexed by goals in $G$.

\begin{algorithm}
    \caption{Backpropagation}
    \label{alg:bp}
    \SetKwFunction{MyBackProp}{BackProp}
    \SetKwProg{Fn}{Function}{:}{}

    \Fn{\MyBackProp{$n, \mathit{fv}, \pi$}}{
        $n'\gets\mathit{Parent}(n)$\;
        $\pi \gets \pi[n.s,n',s]$\;
        $N(\pi,n) \gets N(\pi,n)+1$\;
         \For{each $g\in G$}{
            $q[g]\gets R^\pi_g+\gamma \cdot\mathit{fv[g]}$\;
            $Q(\pi,n,g)\gets Q(\pi,n,g) + \frac{q[g]-Q(\pi,n,g)}{N(\pi,n)}$\;
        }
        \If{$n \neq n_{root}$}{
            \MyBackProp{$\mathit{Parent}(n), q, \pi$}\;
        }
    }
\end{algorithm}

\subsection{The value of a non-GCP HLA}\label{sec:Updating-of-HLAs}

Every non-GCP HLA has a q-value. They are maintained as follows.
Let $\pi_i\to\ldots\to\pi_j$ be the sequence of GCPs that constitutes non-GCP HLA $h[n_i,n_j]$.
Every non-GCP HLA either ends at a leaf or it does not.
That is, either $n_j=n'$ or not.

If $n_j$ is a leaf, then 
\begin{align*}
    Q(h,n_i,g) \doteq &\:R^{\pi_i}_g + \ldots + \gamma^{m-1} R^{\pi_j}_g + Rollout_g(n',m),
\end{align*}
where $m$ is the number of GCPs constituting $h$.
Else if $n_j$ is not a leaf, then
\begin{align*}
    &Q(h,n_i,g) \doteq R^{\pi_i}_g + \ldots + \gamma^{m-1} R^{\pi_j}_g\\
    &\qquad + \gamma^m \max_{h\in HLAs(n_{j+1})}Q(h,n_{j+1},g),\nonumber
\end{align*}
where $n_{j+1}$ is the node at the end of $\pi^j$.

Suppose that $\pi[n,n']$ has just been generated and its q-value propagated back. 
Let the path from the root till leaf node $n'$ be described by the sequence of linked GCPs
\begin{equation}
    \label{eq:GCP-seqence}
\pi_0\to\ldots\to\pi_k.
\end{equation}
Notice that some non-GCP HLAs will be completely or partially constituted by GCPs that are a subsequence of \eqref{eq:GCP-seqence}. Only these HLAs' q-values need to be updated.

\subsection{Desired goal focusing}\label{sec:focus}

An idea is to use a progress-based strategy: Fix the order of the goals in $G$ arbitrarily: $g_1, g_2, \ldots, g_n$. Let $Prog(g,c,t,w,\theta)$ be true iff: the average reward with respect to $g$ experienced over $w$ GCP-steps $c$ is at least $\theta$ or the number of steps is less than $t$. Parameter $t$ is the minimum time the agent should spend learning how to achieve a goal, per attempt.
If $Prog(g_i,c,t,w,\theta)$ becomes false while pursuing $g_i$, then set $c$ to zero, and start pursuing $g_{i+1}$ if $i\neq n$ or start pursuing $g_1$ if $i=n$.
\cite{ckso22} discuss this issue in a section called ``How to Prioritize Goal Selection?''.

\subsection{Executing a task}

\begin{figure}[htbp]
    \centering
    \includegraphics[width=0.5\textwidth]{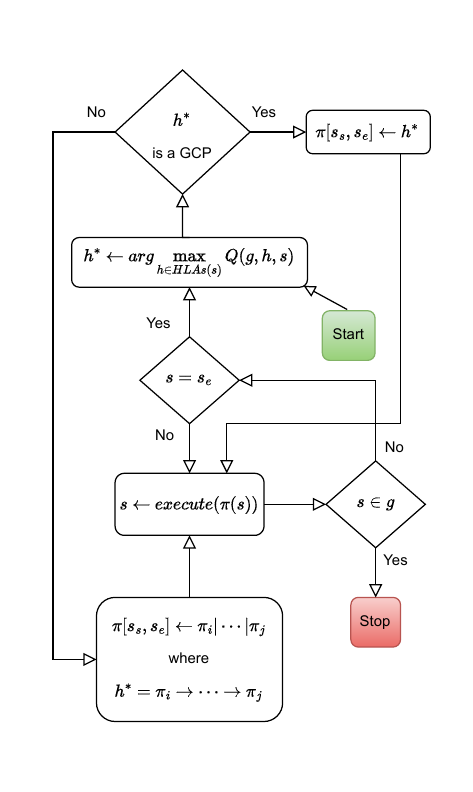}
    \caption{Execution process for a robot to achieve goal $g$ starting in state $s$.}
    \label{fig:execution}
\end{figure}

Once training is finished, the robot can use the generated plan-tree to achieve any particular goal $g\in G$. The execution process is depicted in Figure~\ref{fig:execution}.
Note that $\pi_1 \vert \pi_2$ means that GCPs $\pi_1[s_s^1,s_e^1]$ and $\pi_2[s_s^2,s_e^2]$ are concatenated to form $\pi[s_s^1,s_e^2]$; concatenation is defined if and only $s_e^1=s_s^2$.

\section{\uppercase{Discussion}}\label{sec:Discussion}


\subsection{Learning}

Almost any RL algorithm can be used to learn GCPs, once generated. They might need to be slightly modified to fit the GCRL setting.
For instance, \cite{knst16,zmn23} use DQN \citep{mksrvbgr15}, \cite{ehlsc21} uses PPO \citep{swdrk17}, \cite{yjwlwls22} uses DDPG \citep{lhphetsw16} and \cite{sk23,csl23} use SAC \citep{hzal18}.

\subsection{Representation of Functions}

The three main functions that have to be represented are the contextual goal-conditioned policies (GCPs), GCP values and goal-conditioned Q functions. We propose to approximate these functions with neural networks.

In the following, we assume that every state $s$ is described by a set of features, that is, a feature vector $\mathcal{F}_s$.
Recall that a GCP $\pi[s_s,s_e]$ has context state $s_s$ and behavioral goal state $s_e$. Hence, every GCP can be identified by its context and behavioral goal.

There could be one policy network for all GCPs $\pi[\cdot, \cdot]$ that takes as input three feature vectors: a pair of vectors to identify the GCP and one vector to identify the state for which an action recommendation is sought. The output is a probability distribution over actions $A$; the output layer thus has $\vert A \vert$ nodes. The action to execute is sampled from this distribution.

There could be one policy-value network for all $R_g^\pi$ that takes as input three feature vectors: a pair of vectors to identify the GCP and one vector to identify the state representing the desired goal in $G$.
The output is a single node for a real number.

There could be one q-value network for $Q(h,s,g)$ that takes as input four feature vectors: a pair of vectors to identify the HLA $h$, one vector to identify the state $s$ and one vector to identify the state representing the desired goal $g\in G$.
The output is a single node for a real number.

We could also look at universal value function approximators (UVFAs) for inspiration \citep{shgs15}.

\subsection{Memory}

``Experience replay was first introduced by \cite{l92} and applied in the Deep-Q-Network (DQN) algorithm later on by Mnih et al. (2015).'' \citep{zs17}

Archiving or memorizing a selection of trajectories experienced in a \textit{replay buffer} is employed in most of the algorithms cited in this paper. 
Maintaining such a memory buffer in HGCPP would be useful for periodically updating $ChooseBevGoal(\cdot)$ and the GCP policy network.
We leave the details and integration into the high-level HGCPP algorithm for future work.

Also on the topic of memory, we could generalize the application of generated (remembered) HLAs: Instead of associating each HLA with a plan-tree node, we associate them with a state. In this way, HLAs can be reused from different nodes representing equal or similar states.

\subsection{Opportunism}\label{sec:Opportunism}

Suppose we are attempting to learn $\pi[s,s']$. If no trajectory starting in $s$ reached $s'$ after a given learning-budget, then instead of placing $\pi[s,s']$ in $HLAs(s)$, $\pi[s,s''']$ is placed in $HLAs(s)$, where $s'''$ was reached in at least one experienced trajectory (starting in $s$) and is the best (in terms of $ChooseBevGoal(\cdot)$) of all states reached while attempting to learn $\pi[s,s']$. In the formal algorithm (line 3.d.) $s''$ is either $s'$ or $s'''$.
This idea of relabeling failed attempts as successful is inspired by the Hindsight Experience Replay algorithm \citep{awrsfwmtaz17}.

\bibliographystyle{apalike}
{\small
\bibliography{references}}

\end{document}